%% file: main.tex
\documentclass[letterpaper]{article} %
\usepackage{aaai24}  %
\usepackage{times}  %
\usepackage{helvet}  %
\usepackage{courier}  %
\usepackage[hyphens]{url}  %
\usepackage{graphicx} %
\usepackage{natbib}  %
\usepackage{caption} %
\usepackage{newfloat}
\usepackage{lineno}
\usepackage{listings}
\DeclareCaptionStyle{ruled}{labelfont=normalfont,labelsep=colon,strut=off} %
\title{Temporal Numeric Planning with Patterns}
\author{
    Matteo Cardellini,
    Enrico Giunchiglia
}
\affiliations{
    DIBRIS, Università degli Studi di Genova, Italy\\
     matteo.cardellini@edu.unige.it, enrico.giunchiglia@unige.it
}

\input{macros}
\input{macros_local}

\newtheorem{theorem}{Theorem}

\begin{document}

\maketitle

\begin{abstract}
We consider temporal numeric planning problems $\Pi$ expressed in \pddl2.1 level 3, and show 
how to produce  {\smt} formulas $(i)$ whose models correspond to valid plans of $\Pi$, and $(ii)$ that extend the recently proposed planning with patterns approach from the numeric to the temporal case. We prove the correctness and completeness of the approach and show that it performs very well on 10 domains with required concurrency.
\end{abstract}

\section{Introduction}

We consider temporal numeric planning problems expressed in \pddl2.1 level~3 \cite{Fox_Long_2003}. Differently from the classical case, 
where plans are sequences of instantaneous actions and variables are Boolean, in these problems actions may have a duration, are executed concurrently over time, and can affect Boolean and numeric variables at both the start and  end of their execution.
These two extensions make the problem of finding a valid plan much more difficult --even undecidable in the general case \cite{Helmert_2002,GiganteMMS_2022}-- and  extending state-of-the-art solving techniques from the classical/numeric to the temporal numeric setting is far from  easy. 

In this paper, we extend the recently proposed Symbolic Pattern Planning ({\spp}) approach \cite{CardelliniGM_2024} to handle temporal numeric problems. Specifically, given one such problem $\Pi$ and a bound $n \in \dN^{\ge 0}$, we show how to produce a Satisfiability Modulo Theory ({\smt}) formula \cite{DBLP:series/faia/BarrettSST21} $(i)$ whose models correspond to valid plans of $\Pi$ (correctness),   $(ii)$ which is ensured to be satisfiable for some value of the bound $n$ when $\Pi$ has a valid plan (completeness), and $(iii)$ which is equivalent to  the pattern encoding proposed by \citeauthor{CardelliniGM_2024} when the problem is numeric, i.e., when all the actions are instantaneous.
These  results significantly advance the state-of-the-art, as all symbolic temporal numeric planners are based on the standard encoding with effect and explanatory frame axioms. Given this, we  expect  to obtain also in the temporal setting the substantial improvements  achieved in the 
numeric case, where it was shown $(i)$ that the pattern encoding  {\sl dominates} (i.e., is able
to produce valid plans with a bound $n$ possibly lower and never higher than) both the relaxed-relaxed \re{} encoding~\cite{DBLP:conf/ictai/Balyo13,DBLP:conf/ijcai/BofillEV17}, and the action rolling  $R$ encoding~\cite{Scala_Ramirez_Haslum_Thiebaux_2016_Rolling}; $(ii)$ that both the \re{} and $R$ encodings dominate the standard one, and $(iii)$ that theoretical dominance leads to improved experimental performance, as shown by the  analysis in \cite{CardelliniGM_2024} on the numeric benchmarks of the 2023 International Planning Competition.

To test the effectiveness of our approach, we  compare our planner with all  publicly available temporal planners (both symbolic and based on search) 
on 10 temporal domains with required concurrency \cite{DBLP:conf/ijcai/CushingKMW07}. The results highlight the strong performances of our planner, which achieved the highest coverage (i.e., number of solved problems) in 9 out of  10 domains, while the second-best planner had the highest coverage in  4 domains. Additionally, compared to the other symbolic planners, our system is able to find a valid plan with a lower bound on all the problems. 

The main contributions of this paper are thus:
\begin{enumerate}
    \item We extend the \spp approach to handle temporal numeric problems specified in \pddl2.1.
    \item We prove the correctness and completeness of our     {\smt} encoding.
    \item We conduct an extensive comparative analysis with all available temporal numeric planners, both search-based and symbolic-based, and show that our approach achieves the highest coverage in 9 out of 10 domains.
\end{enumerate}

After the preliminary definitions, we present the basic ideas underlying current standard encodings in {\smt}, followed by the presentation of our approach, the experimental comparative analysis and the conclusions. 
A running example is used to illustrate the features of our encoding.

\section{Preliminaries}\label{sec:prelim}

In PDDL2.1 %
\cite{Fox_Long_2003}
a \textsl{temporal numeric planning problem} is a tuple $\Pi = \langle V_B, V_N,A, I,G\rangle$,
where
\begin{enumerate}
\item 
$V_B$ and $V_N$ are finite sets of {\sl Boolean} and {\sl numeric
variables}, ranging over $\set{\top,\bot}$ and $\Q$ respectively, %
\item 
$I$ is a selected {\sl initial state}, and a {\sl state} is a function mapping each variable to an element in its domain,
\item 
$G$ is a finite set of conditions, called {\sl goals}. A {\sl condition} is either $v = \top$ or  $v=\bot$ or   
$\psi \unrhd 0$, with $v \in V_B$, $\psi$ a linear expression in $V_N$ and $\unrhd \in \{<, \le,=,\ge, >\}$.
\item 
$A$ 
is a finite set of {\sl (instantaneous/snap) actions} and {\sl durative actions}. An {\sl action} $a$ is a pair $\langle\op{pre}(a), \op{eff}(a)\rangle$ %
in which $(i)$ $\op{pre}(a)$ are the {\sl (pre)conditions} of $a$, and $(ii)$ $\op{eff}(a)$ are the  {\sl effects} of $a$ of the form $v \asseq \top$, $v \asseq \bot$, $x \asseq \psi$, with $v \in V_B$, $x \in V_N$ and $\psi$ a linear expression in $V_N$. For each action $a$,  every variable $v \in V_B \cup V_N$ must occur in $\op{eff}(a)$ at most once to the left of the assignment operator ``$\asseq$", and when this happens $v$ is said to be {\sl assigned} by $a$.
A durative action $b$ is a tuple $\langle b^\vdash, b^\vdashv, b^\dashv, [L, U]\rangle$, where $b^\vdash, b^\vdashv, b^\dashv$	are the actions {\sl starting}, {\sl lasting} and {\sl ending} $b$, respectively, and $L,U \in \mathbb{Q}^{> 0}$ are bounds on the  duration $d$ of $b$, $L \le U$. The action $b^\vdashv$ has no effects, and its preconditions $\op{pre}(b^\vdashv)$ must hold throughout the execution of $b$.
From here on, for simplicity, we consider only durative actions, as snap actions can be treated as durative actions without lasting and ending actions, as in \cite{Panjkovic_Micheli_2023}.
\end{enumerate}

Let $\Pi= \langle V_B, V_N,A, I,G\rangle$ be a temporal numeric planning problem. 
A {\sl timed durative action} is a pair $\tuple{t, b}$ with  $t \in \mathbb{Q}^{\ge 0}$ and $b$  a  durative action $\tuple{b^\vdash, b^\dashv,  b^\vdashv, [L, U]}$ in which $[L, U]$ is replaced with a single {\sl duration} value $d \in [L, U]$: 
$t$ (resp. $t+d$) is the time in which  $b^\vdash$ (resp.  $b^\dashv$) is executed. 
A {\sl temporal (numeric) plan} $\pi$ for $\Pi$ is a finite set of timed durative actions. Thus, in $\pi$, multiple snap actions can be executed at the same time, but any two such actions $a$ and $a'$ must be 
non {\sl mutex}, i.e., $a$ must not interfere with $a'$, and vice versa. An action $a$ {\sl does not interfere} with an action $a'$ if for every variable $v$ assigned by $a$
\begin{enumerate}
\item $v$ does not occur in the preconditions of $a'$, and
\item if $v \in V_B$, either $v$ is not assigned by $a'$ or $v \asseq \top \in \op{eff}(a)$ if and only if $v \asseq \top \in \op{eff}(a')$, and
\item if $v \in V_N$ then either $v$ does not occur in the effects of $a'$ or the only occurrences of $v$ in both $a$ and $a'$ are within linear increments of $v$. An expression $v \asseq v + \psi$ is a {\sl linear increment of $v$} if $v$ does not occur in $\psi$.
\end{enumerate}
If $a$ and $a'$ are not in mutex, the order in 
which they are executed in any state $s$ is not relevant, i.e., $res(a,res(a',s)) = res(a',res(a,s))$. The expression $res(a,s)$ is the {\sl result of executing $a$ in state $s$}, which $(i)$ is defined when $s$ satisfies the preconditions of $a$, and $(ii)$ is the state $s'=res(a,s)$ such that for each $v \in V_B\cup V_N$, $s'(v) = s(e)$ if $v \asseq e\in \op{eff}(a)$, and $s'(v) = s(v)$ otherwise. Given a set $A = \set{a_1, \ldots,a_n}$ of pairwise non mutex actions, we write $res(A,s)$ as an abbreviation for $res(a_1,\ldots,res(a_n,s)\ldots)$, order not relevant.

Consider a temporal plan $\pi$.
The execution of $\pi$ induces a sequence of states $s_0, s_1; \ldots; s_m$, each state $s_i$ with an associated time $t_i > t_{i-1}$ at which a non empty set $A_i$ of actions, each starting/ending a durative action in $\pi$, is executed. 
The temporal plan $\pi$ is {\sl valid} if:
\begin{enumerate}
    \item $s_0$ is the initial state, $s_{i+1} = res(A_{i+1},s_i)$ and $s_m$ satisfies the goal formulas, with $i \in [0,m)$;
    \item $\epsilon$-separation:
    for any pair of mutex actions $a \in A_i$ and $a' \in A_j$, $|t_i - t_j| \ge \epsilon > 0$ (and thus $i \neq j$);
    \item no self-overlapping: for any two distinct timed durative actions 
    $\tuple{t,b}$ and $\tuple{t',b}$ with durations $d$ and $d'$ respectively,
    if $t' \ge t$ in $\pi$, then $t' \ge t+d$;
    \item lasting-action:
    for each timed durative action 
    $\tuple{t, \tuple{ b^\vdash, b^\dashv,  b^\vdashv,d}}$ 
    in $\pi$, if $b^\vdash$ and $b^\dashv$ are executed at $t_i=t$ and $t_j=t_i+d$ respectively,
    the preconditions of $b^\vdashv$ are satisfied in  each state $s_i, \ldots, s_{j-1}$.
\end{enumerate}
We thus considered the standard notion of validity used, e.g., in \cite{Fox_Long_2003,RankoohG15,Haslum_pddl_2019,Panjkovic_Micheli_2024}, in which, assuming $V_N = \emptyset$, the problem of deciding the existence of a valid temporal plan is in \textsc{pspace}. Other, more general definitions of plan validity can be given, relaxing the second condition to allow for $|t_i - t_j| > 0$ and/or removing the third condition. With such generalizations, the complexity of deciding the existence of a valid temporal plan, still with $V_N = \emptyset$, can become \textsc{expspace}-complete \cite{Rintanen07} and can even become undecidable \cite{GiganteMMS_2022}.

\section{Standard Encodings in {\smt}}

Several approaches for computing a valid plan of $\Pi$ have been proposed, either based on search (see, e.g., \cite{OPTIC,LPG,TFD}) or on planning as satisfiability 
(see, e.g., \cite{shind04,ShinD05,RankoohG15, Rintanen15, CashmoreFLM16, Rintanen17, CashmoreMZ20, Panjkovic_Micheli_2023, Panjkovic_Micheli_2024}).
We follow the second approach, in which $(i)$ a {\sl bound} or {\sl number of steps} $n$ (initially set to $0$) is fixed, $(ii)$ a corresponding  {\smt} formula is produced, and $(iii)$ a valid plan is returned if the formula is satisfiable, while $n$ is increased and the previous step iterated, otherwise.
In more detail, given a temporal numeric planning problem  $\Pi= \langle V_B, V_N,A, I,G\rangle$ and a value for the bound $n \ge 0$, in the second step, these works:
\begin{enumerate}
    \item 
    Make $n+1$ copies of a set $\mX$ of {\sl state variables}  which includes $V_B \cup V_N$, each copy $\mX_i$ meant to represent the state at the $i$-th step; make
     $n$ copies of a set  $\mA$ of  {\sl (Boolean) durative action variables}  which includes $A$, each copy $\mA_i$ meant to represent the durative actions executed at the $i$-th step; and introduce a set  $\set{t_0, \ldots, t_n}$ of {\sl time variables}, each $t_i$ being the time associated to the $i$-th state $\mX_i$.
    \item 
    Impose proper axioms defining the value of the variables in $\mX_{i+1}$ based on the values of the variables in $\mX_i$, and of the snap actions which are executed in the state $\mX_i$. In particular, these axioms enforce in the state $\mX_{i+1}$ the effects of the actions executed in the state $\mX_i$, and also that no two mutex actions are executed in $\mX_i$.
\end{enumerate}
A similar construction underpins also the standard encoding used  for classical and numeric planning problems. However, in these contexts, the standard encoding is known to underperform compared to the $R$ encoding by \citeauthor{Scala_Ramirez_Haslum_Thiebaux_2016_Rolling} (\citeyear{Scala_Ramirez_Haslum_Thiebaux_2016_Rolling}), the \re{} encoding by \citeauthor{bofill_espasa_villaret_2016} (\citeyear{bofill_espasa_villaret_2016}), and the pattern $\pattern$-encoding by \citeauthor{CardelliniGM_2024} (\citeyear{CardelliniGM_2024}). Indeed, at each step $i \in [0,n)$,
\begin{enumerate}
    \item in the $R$ encoding, each action variable can be ``rolled-up" taking a value in $\natural^{\ge 0}$ representing how many times the action is consecutively executed, 
    \item the \re{} encoding
    allows for the execution of actions in mutex and/or with contradictory effects, and
    \item the $\pattern$-encoding allows for the consecutive execution of actions, even if in mutex  and with contradictory effects.
\end{enumerate}
As a consequence, the $\pattern$-encoding dominates the $\re{}$ and $R$ encodings, which in turn dominate the standard encoding. This dominance usually leads to better performance, as the number of solver calls, along with the number of variables and the encoding size, all increase linearly with the bound~$n$.

To highlight the potential benefits of moving from the standard encoding to the $\pattern$-encoding also in the temporal numeric setting, consider the following simplified version of the bottle example from \cite{ShinD05}.

\begin{iexample}\label{ex:pddl}
There is a set $\{1, \ldots, q\}$ of bottles, the first $p$ of which containing $l_i$ litres of liquid ($i \in [1,p]$),  and the action $\ttt{pr}_{i,j}$ of pouring from the $i$-th bottle (with effects at start) in $[1,p]$ into the $j$-th bottle in $(p,q]$ (with effects at end), one litre every $d_{i,j}$ seconds. In the current encodings, each $\ttt{pr}_{i,j}$ is Boolean  and thus can be executed at most once in between two consecutive states. Further, time variables are associated to the states. For these reasons, with a current encoding $S$, the goal of emptying the bottles in $[1,p]$ needs a number of steps  $n\ge \max_{i=1}^{p} l_i$, how many depending also on the specific $d_{i,j}$ values since each executed $\ttt{pr}_{i,j}$ can start/end at a different time from the others. Further, $S$ needs at least $n=\sum_{i=1}^{p} l_i$ steps when $q = p+1$, due to the conflicting effects of  pouring to a single bottle. 
\end{iexample}

Despite the apparent complexity introduced by the temporal aspects,  \cite{DBLP:conf/ijcai/CushingKMW07} demonstrated that these problems are no more difficult than their  numeric counterparts without the temporal requirements. Indeed, in the above domain each problem admits a solution in which all the durative actions are sequentially executed, one after the other. For this reason, such problems are said to be {\sl without required concurrency}  \cite{DBLP:conf/ijcai/CushingKMW07}, and they can be (more easily) solved by non-temporal planners by $(i)$ replacing  each durative action $b$ with a snap action combining the preconditions and effects of  $b^\vdash, b^\vdashv, b^\dashv$, $(ii)$ finding a sequential solution to the resulting non-temporal problem, and $(iii)$ post-process the found solution to introduce execution times.
We thus consider the following example, whose problems  require concurrency.

\begin{rexample}
    Consider the previous example extended with  $\ttt{nc}_k$ which at start uncaps the bottle $k \in [1,q]$ and then caps it back after $d_k$ seconds. Any problem in which all the bottles are initially capped requires concurrency, since pouring from $i$ to $j$ is possible only if both bottles $i$ and $j$ are uncapped.
This scenario can be modelled in \pddl2.1 with $V_B = \set{c_k \mid k \in [1,q] }$, $V_N = \set{l_k \mid k \in [1,q]}$ and the set of durative actions $A = \set{\ttt{nc}_k \mid k \in [1,q]} \cup \set{\ttt{pr}_{i,j} \mid i \in [1,p], j \in (p,q]}$ whose actions are:
{\small
$$
\begin{array}{c}
    \ttt{pr}_{i,j}^\vdash: \tuple{\set{c_i = \bot, l_i > 0, c_j = \bot},\set{l_i \minuseq 1}}, \\
    \ttt{pr}_{i,j}^\vdashv: \tuple{\set{c_i = \bot, c_j = \bot}, \emptyset}, 
    \ttt{pr}_{i,j}^\dashv: \tuple{\emptyset, \set{l_j \pluseq 1}}, \\
    \hspace{-3mm}\ttt{nc}_{k}^\vdash: \tuple{\set{c_k = \top},\set{c_k \asseq \bot}}, 
    \ttt{nc}_{k}^\dashv: \tuple{\set{c_k = \bot},\set{c_k \asseq \top}}. 
\end{array}
$$}
As customary, $v \pluseq \psi$ (resp. $v \minuseq \psi$) is an abbreviation for $v \asseq v + \psi$ (resp. $v := v - \psi$). With $q=2$ and $p=1$, there are three durative actions $\ttt{pr}_{1,2}$, $\ttt{nc}_{1}$ and $\ttt{nc}_{2}$. Considering the starting/ending actions, $\ttt{pr}_{1,2}^\vdash$ is mutex with $\ttt{nc}_{1}^\vdash$, $\ttt{nc}_{1}^\dashv$.
$\ttt{nc}_{2}^\vdash$, $\ttt{nc}_{2}^\dashv$.  If the bottles are initially capped and  the durations allow to  pour all the litres with just one execution of  $\ttt{nc}_{1}$ and $\ttt{nc}_{2}$), we need a bound
\begin{enumerate}
\item $n = l_1+3$ with the standard encoding (one step for uncapping the bottles, 1 step for starting the first pour action after $\epsilon$ time, $l_1$ steps for pouring the litres and the final step for executing the capping of the bottles),
\item $n = 4$ if we generalize the $R$ encoding since we can roll-up the $\ttt{pr}_{1,2}$ action and collapse the $l_1$ steps into 1,
\item $n = l_1$ if we generalize the $\re{}$ encoding since we can execute all the actions (even the mutex ones) in one step except for the repeated execution of $\ttt{pr}_{1,2}$ (action variables are still Boolean),
\item $n = 1$ if we generalize the $\pattern$-encoding since we can execute all the actions in one step.
\end{enumerate}
If, e.g., $q=4$ and $p=2$ in the standard encoding we need $n = l_1+l_2+3$ steps if the durations of the pour actions forces them to start/end at different times, while we can maintain $n=1$ generalizing the $\pattern$-encoding.
\end{rexample}

\section{Temporal Numeric Planning with Patterns}\label{sec:pattern}

Let $\Pi = \langle V_B, V_N,A, I,G\rangle$ be a temporal numeric planning problem. 
Here, we extend the {\spp} approach to the temporal setting by 
$(i)$ formally defining the notion of pattern $\pattern$ and defining the sets $\mX,\mA^\pattern,\mT^\pattern,\mX'$ of variables used in our encoding;
$(ii)$ extending the definition of rolling to durative actions; 
    $(iii)$ defining the pattern state encoding formula, $T_s^\pattern(\mX,\mA^\pattern,\mX')$, 
    setting the value of each variable in $\mX'$ as a function of $\mX$ and $\mA^\pattern$;
    $(iv)$ defining the pattern time encoding formula, $T_t^\pattern(\mA^\pattern,\mT^\pattern)$, 
    enforcing the desired temporal properties of the actions; and
    $(v)$ proving the correctness and completeness of the presented encoding.
Each point is treated in a separate subsection.
Intuitively, a pattern is a sequence of
starting/ending actions. For each of these actions, the encoding sets $(i)$ an integer specifying how many times the corresponding durative action is executed in sequence, $(ii)$ the conditions for its executability and effects, and $(iii)$ the time at which each durative action sequence has to be started/ended. While durative action sequences that do not interfere might swap order, those that interfere need to maintain the ordering given in the pattern for their starting/ending actions, to ensure executability.

\subsection{Pattern and Language Definition}

A {\sl pattern} is a finite sequence $\pattern = a_1;a_2;\ldots; a_k$ of actions, each starting/ending a durative action in $A$. A pattern is arbitrary, allowing for multiple occurrences of the same action, even consecutively. Each action occurrence in the pattern corresponds to a distinct variable in the encoding, and, given the variable name, we have to be able to uniquely identify
\begin{enumerate}
    \item which durative action it is starting/ending, and
    \item which of the possible multiple occurrences of the action in $\pattern$ we are considering.
\end{enumerate}
For these reasons,
we perform the following two initial steps which do not affect the generality of our approach:
\begin{enumerate}
    \item Whenever in $A$ there are  two distinct durative actions $b_1$ and $b_2$ with $b_1^\vdash = b_2^\vdash$ or $b_1^\vdash = b_2^\dashv$ or $b_1^\dashv = b_2^\dashv$, we break the identity by adding to the preconditions of one of the two actions an always satisfied condition like $0=0$, and
    \item In a pattern $\pattern$, repeated occurrence of an action $a$ are replaced with distinct copies $a'$. Both $a$ and $a'$ are assumed to be starting/ending the same durative action $b$, and, abusing notation, we write, e.g., $a = b^\vdash$ and $a'=b^\vdash$.
    \end{enumerate}
We can therefore take the action in the pattern to be the action variables in our encoding, and we can assume that each action starts/ends exactly one durative action.

Consider a pattern $\pattern = a_1; a_2; \ldots; a_k$, $k \ge 0$. Our encoding is based on the following sets of variables:
\begin{enumerate}
    \item $\mX = V_B \cup V_N$ to represent the initial state;
    \item $\mX'$ containing a {\sl next state variable $x'$} for each state variable $x \in \mX$, used to represent the goal state;
    \item $\mA^\pattern$ consisting of the set of actions in the pattern $\pattern$, each variable $a_i$ ranging over $\dN^{\ge 0}$ and whose value represents the number of times the durative action started/ended by $a_i$ is consecutively executed/rolled up, with $i \in [1,k]$;

    \item $\mT^\pattern$, with $(i)$ a variable $t_i \in \mathbb{Q}^{\ge 0}$ representing the time in which the $i$-th action $a_i$ in $\pattern$ is executed; 
    $(ii)$ if $a_i$ is starting $b$, 
    a variable $d_i \in \mathbb{Q}^{\ge 0}$ representing the time taken by the consecutive execution of $b$ for $p$ times, where $p\ge 0$ is the value assumed by the variable $a_i \in \mA^\pattern$, and $(iii)$ 
for convenience, a variable $t_0=0$ as the initial time.
\end{enumerate}
In the following we keep using $v, w, x$ for state variables, $\psi$ for a linear expression, $a$ for a  (snap) action, $b$ for a durative action, $t$ for a time variable and $d$ for a duration, each symbol possibly decorated with subscripts/superscripts.

\subsection{Rolling Durative Actions}

We start by defining when a durative action $b$ can be rolled up.
Intuitively, $b$ can be consecutively executed more than once when $(i)$ the Boolean effects of its starting/ending actions do not disable the repetition of $b$ given the  preconditions of its starting/lasting/ending actions,  $(ii)$ the numeric effects of $b^\vdash$ and $b^\dashv$ do not interfere between themselves, and $(iii)$ it might be useful to execute $b$ more than once. Formally,  we say that 
$b$ is {\sl eligible for rolling} if the following three conditions are satisfied:
\begin{enumerate}
        \item if $V, V' \in \set{\bot,\top}$, $V \neq V'$, then $(i)$
        $v = V \!\in\! \op{pre}(b^\vdash)$ iff  $v \asseq V \in \op{eff}(b^\dashv)$ or 
        $v \asseq V' \not\in \op{eff}(b^\vdash) \cup \op{eff}(b^\dashv)$, and $(ii)$
        $v = V \!\in\! \op{pre}(b^\vdashv)  \cup \op{pre}(b^\dashv)$ iff $v \asseq V \in \op{eff}(b^\vdash)$ or  
                $v \asseq V' \not\in \op{eff}(b^\vdash) \cup \op{eff}(b^\dashv)$;
    \item if $v \asseq \psi$ is a numeric effect of $b^\vdash$ or $b^\dashv$, then $(i)$ $v$ does not occur in any other effect of $b^\vdash$ or $b^\dashv$, and $(ii)$ either $v$ does not occur in $\psi$ or $v \asseq \psi$ is a linear increment;
    \item $b^\vdash$ or $b^\dashv$ include a linear increment in their effects (this last condition imposed for the usefulness of rolling).
 \end{enumerate}
If $b$ has a duration in $[L, U]$ and is eligible for rolling, 
consecutively executing $b$ for $p\ge 1$ times 
\begin{enumerate}
    \item has a duration in $[p \times L + (p-1) \times \epsilon_b, p \times U + (p-1) \times \epsilon_b]$, where $\epsilon_b = \epsilon$ if $b^\vdash$ and $b^\dashv$ are mutex, and $\epsilon_b =0$ otherwise. 
    Such interval allows for $\epsilon$-separation if $b^\vdash$ and $b^\dashv$ are mutex;%
    \item causes $v$ to get value $(p \times \psi)$ 
if $v \pluseq \psi$ is a linear increment of $b^\vdash$ or $b^\dashv$, while all the other  variables keep the value they get after the first execution of $b$.
\end{enumerate}
Notice that it is assumed that all the consecutive executions of $b$ have the same duration. Indeed, according to the semantics, the duration of $b$ can be arbitrarily fixed as long as each single execution respects the duration constraints, which are part of the domain specification. This assumption does not affect the completeness of our encoding. Should every valid plan require two consecutive executions of $b$ with different durations, we will find a plan when considering a pattern with two or more occurrences of the starting/ending actions of $b$. Indeed, rolling is an optimization, and our procedure is complete even if we rule out rolling by adding the constraint $a \le 1$ for each action $a$.

Then, for each $i \in [0,k]$, the value of a variable $v\in V_B\cup V_N$ after the sequential execution of $a_1; \ldots; a_i$, each action possibly repeated multiple times, is given by $\sigma_i(v)$, inductively defined as $\sigma_0(v) = v$, and, for $i > 0$,
\begin{enumerate}
\item 
if $v$ is not assigned by $a_i$,  $\sigma_i(v) = \sigma_{i-1}(v)$;
    \item 
if $v \asseq \top \in 
      \op{eff}(a_i)$, $\sigma_i(v) = (\sigma_{i-1}(v) \OR a_i > 0)$;
   \item if $v \asseq \bot \in 
      \op{eff}(a_i)$, $\sigma_i(v) = (\sigma_{i-1}(v) \AND a_i = 0)$;
    \item 
if $v \pluseq \psi \in \op{eff}(a_i)$ is a linear increment, 
        $$\sigma_i(v) = \sigma_{i-1}(v) + a_i \times \sigma_{i-1}(\psi),$$
        i.e., the value of $v$ is incremented by $\sigma_{i-1}(\psi)$ a number of times equal to the value assumed by the variable $a_i$;
        \item 
        if $v \asseq \psi \in \op{eff}(a_i)$ is not a linear increment,
        $$\sigma_i(v) = \ite(a_i > 0,
        \sigma_{i-1}(\psi), 
        \sigma_{i-1}(v)),
        $$
        where $\ite(c,
        t, 
        e)$ (for ``\textit{If (c) Then $t$ Else $e$}"
      returns $t$ or $e$ depending on whether $c$ is true or not, and is a standard function in \textsc{smtlib} 
      \cite{BarFT-SMTLIB}.
\end{enumerate}
Above and in the following, for any linear expression $\psi$ and $i \in [0,k]$, $\sigma_i(\psi)$ is the expression obtained by substituting each variable $v \in V_N$ in $\psi$ with $\sigma_i(v)$. 

Given a durative action $b$ eligible for rolling and a state $s$, to determine the maximum number of times that $b$ can be executed consecutively in $s$, we rely on the following Theorem, in which $\psi[p,b^\vdash,q,b^\dashv]$ represents the value of $\psi$ after $p$ and $q$ repetitions of the actions $b^\vdash$ and $b^\dashv$, respectively. Formally, $\psi[p,b^\vdash,q,b^\dashv]$  is the   expression  obtained from $\psi$ by substituting each variable $x$ with 
    \begin{enumerate}
        \item $x + p \times \psi'$ (resp. $x + q \times \psi'$), when $x \pluseq \psi' \in \op{eff}(b^\vdash)$ (resp. $x \pluseq \psi' \in \op{eff}(b^\dashv)$) is a linear increment, and
        \item $\psi''$, when $x \asseq \psi'' \in \op{eff}(b^\vdash) \cup \op{eff}(b^\dashv)$ is not a linear increment.
    \end{enumerate}

\begin{theorem}\label{th:rolling}
    Let $b$ be a durative action eligible for rolling. Let $s$ be a state.  The result of executing $b^\vdash;b^\vdashv;b^\dashv$ consecutively for $p\ge 1$ times in $s$ is defined if and only if 
    for each numeric condition $\psi \unrhd 0$,
    \begin{enumerate}
        \item if $\psi \unrhd 0 \in \op{pre}(b^\vdash)$, $s$ satisfies
        $\psi[0,b^\vdash,0,b^\dashv]\unrhd 0$ (i.e., $\psi \unrhd 0$) and
        $\psi[p-1,b^\vdash,p-1,b^\dashv]\unrhd 0$;
        \item if $\psi \unrhd 0 \in \op{pre}(b^\vdashv) \cup \op{pre}(b^\dashv)$, $s$ satisfies
        $\psi[1,b^\vdash,0,b^\dashv]\unrhd 0$ and
        $\psi[p,b^\vdash,p-1,b^\dashv]\unrhd 0$.
    \end{enumerate}
\end{theorem}
\begin{proof}
The thesis follows from the monotonicity in $p$ of the functions $\psi[p-1,b^\vdash,p-1,b^\dashv]$ and $\psi[p,b^\vdash,p-1,b^\dashv]$ (see  \cite{Scala_Ramirez_Haslum_Thiebaux_2016_Rolling}).   
\end{proof} 

 \begin{rexample}
     For $i \in [1,p]$, $j \in (p,q]$, the pouring action $\ttt{pr}_{i,j}$ is eligible for rolling while both $\ttt{nc}_i$ and  $\ttt{nc}_j$ are not. Action $\ttt{pr}_{i,j}$  
    can be consecutively executed for $l_i$ times in the states in which bottles $i$ and $j$ are uncapped and at least $l_i$ litres are in the $i$-th bottle.
\end{rexample}

\subsection{The Pattern State Encoding}

Let $\pattern = a_1; a_2; \ldots; a_k$, $k \ge 0$, be a pattern.
The pattern state encoding defines the executability conditions of each action and how to compute the value of each variable in $\mX'$ based on the values of the variable in $\mX$ and in $\mA^\pattern$.
Formally, the {\sl pattern state $\pattern$-encoding $T^\pattern_s(\mX,\mA^\pattern,\mX')$ of $\Pi$} is the conjunction of the formulas in the following sets:
\begin{enumerate}

\item $\op{pre}^\pattern(A)$: for each $i \in [1,k]$ and for each 
 $v = \bot$, $w = \top$, $\psi \unrhd 0$  in $\op{pre}(a_i)$:
 \begin{enumerate}
     \item $\neg v$ and $w$ must hold to execute $a_i$:
\begin{equation*}\begin{array}{c}
    a_i > 0 \implies (\neg \sigma_{i-1}(v)
    \AND \sigma_{i-1}(w)), 
\end{array} \end{equation*}
\item
and, if $a_i$ is starting $b$, (i.e., if $a_i = b^\vdash$) 
(Theorem~\ref{th:rolling}):
\begin{equation*}\begin{array}{c}
    a_i > 0 \imp \sigma_{i-1}(\psi[0,b^\vdash,0,b^\dashv]) \unrhd 0, \\    a_i > 1 \imp \sigma_{i-1}(\psi[a_i-1,b^\vdash,a_i-1,b^\dashv]) \unrhd 0, 
\end{array} \end{equation*}  
\item if $a_i$ is ending $b$, (i.e., if $a_i = b^\dashv$)
(Theorem~\ref{th:rolling}, noting that in $\sigma_{i-1}$,  $b^\vdash$ has  been executed $a_i$ times):
\begin{equation*}\begin{array}{c}
    a_i > 0 \imp \sigma_{i-1}(\psi[-a_i+1,b^\vdash,0,b^\dashv]) \unrhd 0, \\
    a_i > 1 \imp \sigma_{i-1}(\psi[0,b^\vdash,a_i-1,b^\dashv]) \unrhd 0. 
\end{array} \end{equation*}  
\end{enumerate}

\item $\op{amo}^\pattern(A)$: for each $i \in [1,k]$, if $a_i$ is starting a durative action which is not eligible for rolling:
  $$      a_i \le 1.$$
\item $\op{frame}^\pattern(V_B\cup V_N)$: for each variable $v\! \in\! V_B$ and $w\! \in\! V_N$:
\begin{equation*}\begin{array}{c}
v' \liff \sigma_k(v), \qquad w' = \sigma_k(w).
\end{array} \end{equation*}     
\end{enumerate}

\begin{rexample}
Assume $p=2$ and $q=4$. Let
\begin{equation}\label{eq:ex-pattern}
\begin{array}{c}
\ttt{nc}_1^\vdash; 
\ttt{nc}_2^\vdash; 
\ttt{nc}_3^\vdash; 
\ttt{nc}_4^\vdash; 
\ttt{pr}_{1,3}^\vdash;
\ttt{pr}_{1,4}^\vdash;
\ttt{pr}_{2,3}^\vdash;
\ttt{pr}_{2,4}^\vdash; \\
\ttt{nc}_1^\dashv; 
\ttt{nc}_2^\dashv; 
\ttt{nc}_3^\dashv; 
\ttt{nc}_4^\dashv; 
\ttt{pr}_{1,3}^\dashv;
\ttt{pr}_{1,4}^\dashv;
\ttt{pr}_{2,3}^\dashv;
\ttt{pr}_{2,4}^\dashv.
\end{array}
\end{equation}
be the fixed pattern $\pattern$. Assume $i\in [1,2]$, $j \in [3,4]$, $k\in[1,4]$. 
The pattern state encoding entails $(nc^\vdash_k \le 1)$ since the durative action $nc$ is not eligible for rolling, and
{\small
\begin{equation*}
\begin{array}{c}
    \ttt{nc}^\vdash_{i} > 0 \imp c_i, \ \
    \ttt{nc}^\dashv_{i} > 0 \imp \neg (c_i \wedge \ttt{nc}^\vdash_{i} = 0), \\
    \ttt{pr}^\vdash_{i,j} > 0 \imp (\neg (c_i \wedge \ttt{nc}_i^\vdash = 0) \wedge \neg (c_j \wedge \ttt{nc}_j^\vdash = 0)), \\
    \ttt{pr}^\vdash_{i,3} > 0 \imp  l_i > 0, \qquad
    \ttt{pr}^\vdash_{i,4} > 0 \imp l_i - \ttt{pr}^\vdash_{i,3} > 0, \\
    \ttt{pr}^\vdash_{i,3} > 1 \imp  \ttt{pr}^\vdash_{i,3} < l_i, \qquad
    \ttt{pr}^\vdash_{i,4} > 1 \imp  \ttt{pr}^\vdash_{i,4} < l_i - \ttt{pr}^\vdash_{i,3}, \\
c'_k \equiv (c_k \wedge \ttt{nc}_k^\vdash = 0) \vee \ttt{nc}_k^\dashv > 0, \\
l'_i = l_i - \ttt{pr}^\vdash_{i,3} - \ttt{pr}^\vdash_{i,4}, \qquad
l'_j = l_j + \ttt{pr}^\dashv_{1,j} + \ttt{pr}^\dashv_{2,j}.
\end{array}
\end{equation*}
}
The first four lines define the preconditions for executing each action, and the last two specify the frame axioms.
\end{rexample}

As the frame axioms in the example make clear, the $\pattern$-encoding allows in the single state transition from $\mX$ to $\mX'$ $(i)$ the multiple consecutive execution of the same action, as in the rolled-up $R$ encoding \cite{Scala_Ramirez_Haslum_Thiebaux_2016_Rolling}, and $(ii)$ the combination of multiple even contradictory effects on a same variable by different actions, as in the \re{} encoding \cite{bofill_espasa_villaret_2016}.

\subsection{The Pattern Time Encoding}

Let $\pattern = a_1; a_2; \ldots; a_k$, $k \ge 0$, be a pattern.
The pattern time $\pattern$-encoding associates to each action $a_i$ in $\pattern$ a starting time $t_i$ and duration $d_i$, which are both set to 0 when $a_i$ is not executed, i.e., when $a_i=0$. In defining the constraints for $t_i$ and $d_i$ they have to respect the semantics of temporal planning problems and also the causal relations between the actions in the pattern and exploited in the pattern state $\pattern$-encoding. Consider for instance two actions $a_i$ and $a_j$ in $\pattern$ with $i < j$, $a_i > 0$ and $a_j > 0$. We surely have to guarantee that $t_i < t_j$ if $a_i$ and $a_j$ are in mutex: 
the formulas checking that the preconditions of $a_j$ (resp. $a_i$) are satisfied, take into account that $a_i$ (resp. $a_j$) has been (resp. has not been) executed {\em before} $a_j$ (resp. $a_i$). Even further,  we have to impose that $t_i + \epsilon \le t_j$ for the $\epsilon$-separation rule. If, on the other hand, $a_i$ and $a_j$ are not in mutex, then it is not necessary to guarantee $t_i < t_j$ unless $a_j$ is ending the durative action started by $a_i$ or because of the lasting action of the durative action started by $a_j$. As an example of the impact of the lasting action on the encoding, assume $a_j$ is starting action $b$. Then, it may be the case $a_i$ is not in mutex with $a_j$ but it is in mutex with the lasting action $b^\vdashv$ of $b$. Hence,  the formulas checking the executability  of $b^\vdashv$ encode that $a_i$ precedes $a_j$ in the pattern, and consequently we will have to guarantee $t_i < t_j$.

Given the above, the {\sl pattern time $\pattern$-encoding $T^\pattern_t(\mA^\pattern,\mT^\pattern)$ of $\pattern$} is the conjunction of 
$(t_0 = 0)$ and 
the following formulas:
\begin{enumerate}
    \item $\op{dur}^\pattern(A)$: for each durative action $\langle b^\vdash, b^\dashv,  b^\vdashv, [L, U]\rangle \in A$ and for each action $a_i = b^\vdash$ and $a_j = b^\dashv$ in $\pattern$: 
    $$
    \begin{array}{c}
    a_i > 0 \imp t_i \ge t_0 + \epsilon, \\
    a_i = 0 \imp t_i = t_0 \wedge d_i = 0, \ \ a_j = 0 \imp t_j = t_0, \\
    a_i > 0 \imp 
    a_i \times (L + \epsilon_b)  \le d_i + \epsilon_b \le a_i \times (U + \epsilon_b).
    \end{array}
    $$
    The last formula guarantees also $\epsilon$-separation when $b$ is consecutively executed, and $b^\vdash$ and $b^\dashv$ are in mutex. 
    \item 
    $\op{start}$-$\op{end}^\pattern(A)$: for each durative action $b$, each starting action $a_i = b^\vdash$ (resp. ending action $a_j = b^\dashv$) in $\pattern$ must have a matching ending (resp. starting) action:
$$
\begin{array}{c}
    a_i > 0 \implies \bigor_{j \in E_i} (a_i = a_j \AND t_j = t_i + d_i), \\
    a_j > 0 \implies \bigor_{i \in S_j} (a_i = a_j \AND t_j = t_i + d_i),
\end{array}$$
    where $E_i = \{j \in (i,k] \mid a_i = b^\vdash, a_j = b^\dashv\}$, and $S_j = \{i \in [1,j) \mid a_j = b^\dashv, a_i = b^\vdash\}$.  
    \item $\op{epsilon}^\pattern(A)$: every two actions $a_i$ and $a_j$ in $\pattern$ with $j < i$ are $\epsilon$-separated if they are mutex or different copies of the same action:
    $$
    a_i > 0 \imp (t_i \ge t_j + \epsilon).
    $$
    Further, for every   two actions $a_i$ and $a_j$ starting respectively $b$ and $b'$, if the starting or ending action of $b$ is mutex with the starting or ending action of $b'$:
    $$
    \setlength{\arraycolsep}{0pt}
    \begin{array}[t]{rl} 
a_i > 1 \imp (& t_i \ge t_j + d_j \vee t_j \ge t_i + d_i \vee \\
& a_j = 1 \wedge t_i \ge t_j \wedge t_i + d_i \le t_j + d_j).
    \end{array}$$
    This formula ensures that the start/end actions of $b'$
  are not  executed during the multiple consecutive executions  of $b$, thereby guaranteeing $\epsilon$-separation.
    \item $\op{noOverlap}^\pattern(A)$: for each durative action $b$, each starting action $a_i=b^\vdash$  in $\pattern$ can be executed only after the previous executions of $b$ ended:
$$
\begin{array}{c}
    a_i > 0 \implies \bigand_{j \in B_i} (t_i \ge t_j + d_j), \\
\end{array}$$    
    where $B_i = \set{j \in [1,i) \mid a_i = b^\vdash, a_j = b^\vdash}$.
    \item $\op{lasting}^\pattern(A)$: for each durative action $b$ with $\op{pre}(b^\vdashv) \neq \emptyset$, and
    for each action $a_i = b^\vdash$ in $\pattern$:
    \begin{enumerate}
        \item The preconditions of $b^\vdashv$ must be satisfied in each (consecutive) execution of $b$, i.e., for each $v = \bot$, $w = \top$, $\psi \unrhd 0$  in $\op{pre}(b^\vdashv)$ (Theorem~\ref{th:rolling}):
        \begin{equation*}
        \begin{array}{c}
\!\!\!\!            a_i > 0 \imp \neg\sigma_i(v) \wedge \sigma_i(w) \wedge    \sigma_{i-1}(\psi[1,b^\vdash,0,b^\dashv]) \unrhd 0, \\
            a_i > 1 \imp
\sigma_{i-1}(\psi[a_i,b^\vdash,a_i-1,b^\dashv]) \unrhd 0.
    \end{array}
    \end{equation*}
    \item For each action $a_j$ in $\pattern$ mutex with  $b^\vdashv$, 
    \begin{enumerate}
        \item if $j < i$, then $a_j$ cannot be executed after $a_i$:
            $$a_i > 0 \imp t_i \ge t_j,$$
    and,  when $a_j$ is a starting action, also:
    $$
    \begin{array}{c}
    a_i > 0 \wedge a_j > 1 \imp t_i \ge t_j + d_j.
    \end{array}
    $$
    These formulas ensure that $b$ does not start until all executions of $a_j$ happened.
    \item 
    if $j > i$ and $a_j$ is executed before $b$ ends,
    then $(i)$ no rolling takes place: 
    $$t_0 + \epsilon \le t_j < t_i + d_i \imp a_i \le 1 \wedge a_j \le 1,$$
    and $(ii)$
    $a_j$ has to maintain the preconditions of $b^\vdashv$, i.e.,
    for each 
     $v = \bot$, $w = \top$, $\psi \unrhd 0$  in $\op{pre}(b^\vdashv)$:
$$
    t_0 + \epsilon \le t_j < t_i + d_i \imp
    \neg\sigma_j(v) \wedge \sigma_j(w) \wedge \sigma_j(\psi) \unrhd 0.
    $$
    \end{enumerate}
    \end{enumerate}
\end{enumerate}

\begin{rexample}
For $\ttt{pr}^\vdash_{i,j}$ (resp. $\ttt{nc}^\vdash_{k}$), let $t^\vdash_{i,j}$ (resp. $t^\vdash_{k})$ be the associated time variable, and analogously for the  ending actions. 
If we further assume that when executed, the durations $d_{i,j}$ and $d_k$ 
 of $\ttt{pr}_{i,j}$ and $\ttt{nc}_k$ are 1 and 5 respectively, the temporal pattern encoding entails:
\begin{equation*}
\begin{array}{c}
    \ttt{nc}^\vdash_{k} = 0 \imp d_k = 0, 
    \ttt{nc}^\vdash_{k} = 1 \imp d_k = 5, 
    \ttt{pr}^\vdash_{i,j} = d_{i,j}, \\
    \ttt{nc}^\vdash_{k} = \ttt{nc}^\dashv_{k},\ \
    \ttt{pr}^\vdash_{i,j} = \ttt{pr}^\dashv_{i,j},\quad
    \neg (t^\vdash_{i,j} \le t^\dashv_i < t^\vdash_{i,j} + d_{i,j}), \\
    \ttt{pr}^\vdash_{i,j} > 0 \wedge \ttt{nc}^\vdash_{i} = 1 \imp t^\vdash_{i,j} \ge t^\vdash_{i} + \epsilon.
\end{array}
\end{equation*}
The formulas in the 3 lines respectively say that $(i)$  uncapping a bottle takes 5s and pouring $p$ litres takes $p$ seconds, $(ii)$  any started durative action has to be ended and it is not possible to cap a bottle while pouring from it, and $(iii)$
 we can start pouring from a bottle after $\epsilon$ time since we uncapped it.
Similar facts hold for the destination bottles.
\end{rexample}

\input{assets/table}

\subsection{Correctness and Completeness Results}

Let $\pattern = a_1; a_2; \ldots; a_k$, $k \ge 0$, be a pattern.
Though the pattern $\pattern$ can correspond to any sequence of starting/ending actions of a durative action in $A$, it is clear that it is pointless to have $(i)$ an ending action $b^\dashv$ without the starting action $b^\vdash$ before $b^\dashv$ in $\pattern$; similarly $(ii)$ a starting action $b^\vdash$ which is not followed by  the ending action $b^\dashv$, and $(iii)$ two consecutive occurrences of the same starting (ending) action in the pattern. In such cases, the pattern can be safely simplified by eliminating such actions. On the other hand, it makes sense to consider patterns with non consecutive occurrences of the same starting/ending action.  Assuming $b_1$ and $b_2$ are two durative actions with $b_1^\vdash/b_1^\dashv$ mutex with $b_2^\vdash$, it might be useful to have a pattern including $b_1^\vdash;b_1^\dashv;b_2^\vdash;b_1^\vdash;b_1^\dashv$ to allow two executions of $b_1$, or $b_1$ to start/end before/after $b_2$ starts. 
No matter how $\pattern$ is defined,
 the {\sl $\pattern$-encoding $\Pi^\pattern$ of $\Pi$ (with bound $1$)} is correct, where
{\small\begin{equation}
    \label{eq:enc_pattern_n}
\Pi^\pattern = 
I(\mX) \AND T^\pattern_s(\mX,\mA^\pattern,\mX') \AND 
T^\pattern_t(\mA^\pattern,\mT^\pattern) \AND G(\mX'),
\end{equation}
}in which  $I(\mX)$ and $G(\mX')$ are formulas encoding the 
initial state and the goal conditions.
To any model $\mu$ of $\Pi^\pattern$ we associate the valid temporal plan $\pi$ whose durative actions are started by the actions $a_i$ in $\pattern$ with $\mu(a_i) > 0$. Specifically, if $a_i = b^\vdash$, in $\pi$ we have $\mu(a_i)$ consecutive executions of $b$, i.e., one timed durative actions $\tuple{t,\tuple{b^\vdash,b^\vdashv,b^\dashv,d}}$ for each value of $p \in [0,\mu(a_i))$. The $(p+1)$-th execution of $b$ happens at the time $t$ and has duration $d$ such that
$$
\begin{array}c
    t=\mu(t_i) + p \times (d+\epsilon_b),\ (d+\epsilon_b)\times\mu(a_i) = \mu(d_i)+\epsilon_b.
\end{array}
$$

Completeness is guaranteed once we ensure that  the sequence $\indpi$ of the starting/ending actions of a valid temporal plan $\pi$, listed according to their execution times, is a subsequence of the pattern used in the encoding. This can be achieved by starting with a complete pattern, and then repeatedly chaining it till $\Pi^\pattern$ becomes satisfiable.
Formally, a pattern $\pattern$ is {\sl complete} if for each durative action $b \in A$, $b^\vdash$ and $b^\dashv$ occur in $\pattern$. Then, we define $\pattern^n$ to be the sequence of actions obtained concatenating $\pattern$ for $n \ge 1$ times. Finally,
$\Pi^\pattern_n$ is the {\sl pattern $\pattern$-encoding of $\Pi$ with bound $n$}, obtained from (\ref{eq:enc_pattern_n}) by considering $\pattern^n$ as the pattern $\pattern$.

\begin{theorem} \label{th:compl}
Let $\Pi$ be a temporal numeric planning problem. Let $\pattern$ be a pattern. %
Any model of $\Pi^\pattern$ corresponds to a valid temporal plan of $\Pi$ (correctness). If $\Pi$ admits a valid temporal plan and $\pattern$ is complete, then for some $n\ge 0$, $\Pi^\pattern_n$ is satisfiable (completeness).
\end{theorem}

\begin{proofsketch}
    Correctness: Let $\mu$ be a model of $\Pi^\pattern$ and $\pi$ its associated plan.  The $\epsilon$-separation axioms ensure that the relative order between mutex actions in $\pi$ and in $\pattern$ is the same. The pattern state encoding ensures that  executing sequentially the actions in $\pi$ starting from $I$  leads to a goal state. The axioms in the pattern time encoding are a logical formulation of the corresponding properties for the validity of $\pi$. Completeness: Let $\pi$ be a valid temporal plan with $n$ durative actions. Let $\pattern_\pi$ be  the pattern consisting of the starting and ending actions in $\pi$ listed according to their execution times. The formula $\Pi^{\pattern_\pi}$ is satisfied by the model $\mu$ whose associated plan is $\pi$. For any complete pattern $\pattern$, $\pattern_\pi$ is a subsequence of $\pattern^{2\times n}$ and  $\Pi^\pattern_{2\times n}$ can be satisfied by extending $\mu$ to assign 0 to all the action variables not in $\pattern_\pi$. 
\end{proofsketch}

\noindent Notice that when two actions $a$ and $a'$ are not in mutex and one is not the starting/ending action of the other, the pattern does not lead to an ordering on their execution times. For this reason, we may find a valid plan $\pi$ for $\Pi$ even before $\pattern^n$ becomes a supersequence of $\indpi$, $\indpi$ defined as above.

\begin{rexample}
Assume all $q \ge 2\times p$ bottles are initially capped and that the bottles in $[1,p]$ contain  $< d_k = 5$ litres. Then, $\Pi^\pattern$ is satisfiable and a valid plan is found with one call to the {\smt} solver. Notice that in the pattern (\ref{eq:ex-pattern}), the ending action $\ttt{pr}^\dashv_{i,j}$ of the pouring actions are after the ending action $\ttt{nc}^\dashv_{k}$ that caps the bottle. However, such two actions are not in mutex and our pattern time encoding does not enforce $t^\dashv_{i,j} > t^\dashv_{k}$, making it possible to solve the problem with a bound $n=1$. On the other hand,
if one bottle contains 5 litres, $\Pi^\pattern$ is unsatisfiable because of $\epsilon$-separation between the actions of uncapping and pouring from it, making it impossible to pour 5 times  before the bottle is capped again. This problem is solved having $\pattern^n$ with $n=2$.
More complex scenarios may require $\pattern^n$ with higher values for $n$. 
\end{rexample}

\section{Experimental Results}\label{sec:exp}

Table \ref{tab:experiments} presents the experimental analysis on the \textsc{Cushing} domain (the only domain with required concurrency in the last International Planning Competition ({\ipc}) with a temporal track \cite{ipc2018}),
all the domains and problems presented in \cite{Panjkovic_Micheli_2023} (last five),  and four new domains covering different types of required concurrency specified in \cite{DBLP:conf/ijcai/CushingKMW07}.
The first new domain, \textsc{Pour}, is similar to the motivating example of this paper. 
\textsc{Shake} allows emptying a bottle by shaking it while uncapped. \textsc{Pack} calls for concurrently pairing two bottles together to be packed. The domain \textsc{Bottles} puts together all the actions and characteristics of the three aforementioned domains. Of these 10 domains, only \textsc{Pour} and \textsc{Bottles}, contain actions eligible for rolling.

The analysis compares our system \pattyt implemented by modifying the planner \patty \cite{CardelliniGM_2024} and using the {\smt}-solver \ttt{Z3} v4.8.7 \cite{de2008z3}; the symbolic planners \textsc{AnmlSMT} (which corresponds to $\text{ANML}^{\text{OMT}}_{\text{INC}}(\text{OMSAT})$ in \cite{Panjkovic_Micheli_2023}) and \textsc{ITSat} \cite{RankoohG15}; and the search-based planners \textsc{Optic} \cite{OPTIC}, \textsc{LPG} \cite{LPG} and \textsc{TemporalFastDownward} (\textsc{TFD}) \cite{TFD}. \textsc{AnmlSMT} and \textsc{Optic} have been set in order to return the first valid plan they find. To use  \textsc{AnmlSMT}, we manually converted the domains in \pddl2.1 to the \textsc{ANML} language \cite{smith2008anml}. The experiments have been run using the same settings used in the Numeric/Agile Track  of the last {\ipc}, with 20 problems per domain and a time limit of $5$ minutes. Analyses have been run on an Intel Xeon Platinum $8000$ $3.1$GHz with 8 GB of RAM.
In the table we show: the percentage of solved instances (Coverage); the average time to find a solution, counting the time limit when the solution could not be found (Time); the average bound at which the solutions were found, computed on the problems solved by all the symbolic planners able to solve at least one problem in the domain (Bound). The value of the bound coincides with the number of calls to the {\smt} solver.
Each pattern $\pattern$ is computed only once using the Asymptotic Relaxed Planning Graph, introduced in \cite{Scala_Haslum_Thiebaux_Ramirez_2016_AIBR} and already used in \cite{CardelliniGM_2024}.
\footnote{\pattyt{} and the \pddl2.1 and ANML encoding of the new domains are available at \url{https://github.com/matteocarde/patty}
.}

From the table, as expected \pattyt finds a solution with a bound always lower than the ones needed by the other symbolic planners. This allows \pattyt to have the highest coverage in 9 out of 10 domains (compared to the value 4 for the second best). The Painter domain is the only one where \pattyt has a lower coverage than \textsc{AnmlSMT}. \textsc{AnmlSMT} is a symbolic planner exploiting the standard encoding. Although it requires a higher bound to find a valid plan also in Painter, \textsc{AnmlSMT} encoding has 2490 mostly Boolean variables (action and most state variables are Boolean), while our encoding has 2058 mostly numeric variables (the only Boolean variables are in $\mX$ and $\mX'$). In the other domains, the ratio between the number of variables used by \textsc{AnmlSMT} and \pattyt is 0.16 on average, which provides an  explanation of \pattyt's highest coverage and better performance on 9/10 and 6/10 domains, respectively.
Overall, \pattyt is able to solve 157 out of the 200 considered problems, compared to the 98 of the second best.

\section{Conclusion}\label{sec:concl}

We extended the {\spp} approach proposed in  \cite{CardelliniGM_2024} to the temporal numeric setting. 
We proved its correctness and completeness, and showed its  benefits on various domains with required concurrency. As expected, all the problems have been solved by \pattyt with a bound lower than the one needed by the other planners based on planning as satisfiability. 

\section{Acknowledgments}
Enrico Giunchiglia acknowledges the financial support from PNRR MUR Project PE0000013 
 FAIR “Future Artificial Intelligence Research”, funded by the European Union – NextGenerationEU, CUP J33C24000420007, and from Project PE00000014 “SEcurity and RIghts in the CyberSpace”, CUP D33C22001300002.

\bibliography{biblio}

\end{document}

%% file: macros.tex
\usepackage{xspace}
\usepackage{amsmath}
\usepackage{amsfonts}
\usepackage{amssymb}
\usepackage{comment}
\usepackage{bm}
\usepackage{amsthm}
\usepackage{todonotes}

\newcommand{\set}[1]{\ensuremath{\{#1\}}\xspace}
\newcommand{\tuple}[1]{\ensuremath{\langle #1 \rangle}\xspace}

\newcommand{\op}[1]{\mathrm{#1}}

\newcommand{\ttt}{\texttt}

\newcommand{\pluseq}{\mathrel{+}=}
\newcommand{\minuseq}{\mathrel{-}=}

\newcommand{\OR}{\vee}
\newcommand{\AND}{\wedge}

\newcommand{\mA}{\mathcal{A}\xspace}

\newcommand{\mT}{\mathcal{T}\xspace}

\newcommand{\mX}{\mathcal{X}\xspace}

\newcommand{\dN}{\mathbb{N}\xspace}

\renewcommand{\natural}{\ensuremath{\mathbb{N}}\xspace}

\usepackage{pifont}%

\newcommand{\pddl}{\textsc{pddl}\xspace}

\renewcommand{\implies}{\rightarrow}
\newcommand{\imp}{\implies}
\newcommand{\liff}{\leftrightarrow}

\newcommand{\patty}{\textsc{patty}\xspace}

\newcommand{\pattern}{{\prec\xspace}}

\newcommand{\bigand}{\bigwedge}
\newcommand{\bigor}{\bigvee}

\usepackage{ stmaryrd }

\newcommand{\asseq}{:=}

\newcommand{\vdashv}{{\vdash\mathrel{\mkern-9mu}\dashv}}

\newcommand{\smt}{\textsc{smt}\xspace}

\newcommand{\spp}{\textsc{spp}\xspace}

\newcommand{\ipc}{\textsc{ipc}\xspace}
\newcommand{\ite}{\textsc{ite}\xspace}

\DeclareFontFamily{U}{mathb}{\hyphenchar\font45}
\DeclareFontShape{U}{mathb}{m}{n}{
<-6> mathb5 <6-7> mathb6 <7-8> mathb7
<8-9> mathb8 <9-10> mathb9
<10-12> mathb10 <12-> mathb12
}{}
\DeclareSymbolFont{mathb}{U}{mathb}{m}{n}
\DeclareMathSymbol{\cespattern}{\mathrel}{mathb}{"CE}

%% file: macros_local.tex
\newcommand{\notadded}[1]{}

\newcommand{\Q}{\mathbb{Q}}

\newtheorem*{iexample}{Example}
\newtheorem*{rexample}{Example (cont'd)}

\renewcommand{\pattern}{{\prec}}

\newcommand{\re}{{\ensuremath{R^2\exists}}}

\newcommand{\indpi}{\pi^{\shortdownarrow}}

\newcommand{\pattyt}{$\textsc{patty}_\textsc{t}$\xspace}
\newenvironment{proofsketch}{%
  \proof}{\endproof}

%% file: assets/table.tex
\begin{table*}[tb]
            \centering
            \resizebox{\textwidth}{!}{
\Huge{
\begin{tabular}{|l||cccccc||cccccc||ccc||}
\hline
 & \multicolumn{6}{c||}{Coverage (\%)}&\multicolumn{6}{c||}{Time (s)}&\multicolumn{3}{c||}{Bound (Common)}\\
Domain & \pattyt&$\textsc{AnmlSMT}$&$\textsc{ITSat}$&$\textsc{LPG}$&$\textsc{Optic}$&$\textsc{TFD}$&\pattyt&$\textsc{AnmlSMT}$&$\textsc{ITSat}$&$\textsc{LPG}$&$\textsc{Optic}$&$\textsc{TFD}$&\pattyt&$\textsc{AnmlSMT}$&$\textsc{ITSat}$\\
\hline
\textit{Temporal}&\textbf{9}&\textbf{4}&\textbf{2}&\textbf{1}&\textbf{4}&\textbf{0}&\textbf{6}&\textbf{1}&\textbf{0}&\textbf{0}&\textbf{3}&\textbf{0}&\textbf{10}&\textbf{0}&\textbf{0}\\\hline
\textsc{Cushing}&\textbf{100.0}&30.0&-&-&\textbf{100.0}&10.0&\textbf{1.70}&235.35&-&-&3.12&270.02&\textbf{3.00}&11.33&-\\
\textsc{Pour}&\textbf{95.0}&5.0&-&-&-&-&\textbf{46.51}&285.96&-&-&-&-&\textbf{2.00}&15.00&-\\
\textsc{Shake}&\textbf{100.0}&50.0&-&-&-&-&\textbf{1.11}&155.15&-&-&-&-&\textbf{2.00}&9.50&-\\
\textsc{Pack}&\textbf{60.0}&5.0&-&-&-&-&\textbf{154.72}&285.00&-&-&-&-&\textbf{1.00}&6.00&-\\
\textsc{Bottles}&\textbf{10.0}&5.0&-&-&-&-&\textbf{284.28}&286.36&-&-&-&-&\textbf{7.00}&18.00&-\\
\textsc{Majsp}&\textbf{85.0}&50.0&-&-&-&-&\textbf{90.54}&154.02&-&-&-&-&\textbf{8.40}&15.00&-\\
\textsc{MatchAC}&\textbf{100.0}&\textbf{100.0}&\textbf{100.0}&-&\textbf{100.0}&-&2.20&0.46&0.71&-&\textbf{0.01}&-&\textbf{3.85}&10.00&4.00\\
\textsc{MatchMS}&\textbf{100.0}&\textbf{100.0}&\textbf{100.0}&-&\textbf{100.0}&-&1.22&0.43&0.68&-&\textbf{0.01}&-&\textbf{3.60}&10.00&4.00\\
\textsc{Oversub}&\textbf{100.0}&\textbf{100.0}&-&\textbf{100.0}&\textbf{100.0}&-&1.02&0.05&-&0.08&\textbf{0.01}&-&\textbf{1.00}&4.00&-\\
\textsc{Painter}&35.0&\textbf{45.0}&-&-&10.0&-&211.69&\textbf{194.67}&-&-&270.03&-&\textbf{2.40}&16.80&-
\\\hline

        \end{tabular}}}
        \caption{Comparative analysis.
        A ``-" indicates that no result was obtained in our 300s time limit,
either due to a timeout or an issue with the planner. %
        The best results are in bold.}
        \label{tab:experiments}
        \vspace*{-4mm}
        \end{table*}